\documentclass[journal,onecolumn,table]{IEEEtran}
\ifCLASSINFOpdf
\else
\fi
\usepackage{tikz,lipsum,lmodern}
\usepackage[most]{tcolorbox}

\usepackage{multirow}
\usepackage{longtable}
\usepackage{graphicx}
\usepackage{tabularx}
\usepackage{xcolor}
\usepackage{url}

\definecolor{Gray}{gray}{0.85}
\definecolor{LightCyan}{rgb}{0.88,1,1}

\newcolumntype{a}{>{\columncolor{Gray}}c}
\newcolumntype{b}{>{\columncolor{white}}c}

\usepackage{refstyle}
\begin{document}
%
\title{TweetBERT: A Pretrained Language Representation Model for Twitter Text Analysis}
%
%
%

\author{Mohiuddin Md Abdul Qudar and~Vijay Mago
\thanks{Mohiuddin Md Abdul Qudar and Vijay Mago are with Lakehead University.}}

\maketitle

\begin{abstract}
Twitter is a well-known microblogging social site where users express their views and opinions in real-time. As a result, tweets tend to contain valuable information. With the advancements of deep learning in the domain of natural language processing, extracting meaningful information from tweets has become a growing interest among natural language researchers. Applying existing language representation models to extract information from Twitter does not often produce good results. Moreover, there is no existing language representation models for text analysis specific to the social media domain. Hence, in this article, we introduce two TweetBERT models, which are domain specific language presentation models, pre-trained on millions of tweets. We show that the TweetBERT models significantly outperform the traditional BERT models in Twitter text mining tasks by more than 7\% on each Twitter dataset. We also provide an extensive analysis by evaluating seven BERT models on 31 different datasets. Our results validate our hypothesis that continuously training language models on twitter corpus help performance with Twitter 

\end{abstract}


%
\IEEEpeerreviewmaketitle

\section{Introduction}
Twitter is a popular social networking platform where users tend to express themselves and share their information in real time \cite{castillo2011information}. As a result, text from Twitter is widely studied by natural language researchers and social scientists. The users tend to write texts that are very colloquial and casual \cite{castillo2011information}. The text is are written in a completely different way than traditional writings, primarily due to a restriction in their length. However, their contents are powerful enough to start a movement all over the world or to detect a pandemic in the early stages \cite{go2009twitter}. Hence, usage and style of language is extensively studied. The users, in any social media platform, are more comfortable in discussing their views and perspectives in an informal manner, usually following very little or no grammatical rules \cite{giatsoglou2017sentiment} \cite{wang2007topical}. This is often seen in Twitter due to its character limit of only 280 characters\footnote{https://about.twitter.com/} per tweet. Using existing language representation models, such as BERT (Bidirectional Encoder Representations from Transformers) \cite{devlin2018bert} or AlBERT \cite{lan2019albert}, to evaluate such texts is a challenge. As a result, a need for a language model specific to social media domain arises. Deep neural network models have contributed significantly to many recent advancements in natural language processing (NLP), especially with the introduction of BERT. BERT and BioBERT \cite{alsentzer2019publicly} have considerably improved performance on datasets, in the general domain and biomedical domain, respectively. State-of-art research indicates that when unsupervised models are pre-trained on large corpora, they perform significantly better in NLP tasks. However, language models, such as BioBERT, cannot achieve high performance on domains like social media corpora. This is mainly due to the fact that these models are trained on other domain corpora and the language in social media is irregular and mostly informal. To address this need, in this article, TweetBERT is introduced, which is a language representation model that has been pre-trained on a large number of English tweets, for conducting Twitter text analysis. Experimental results show that TweetBERT outperformed previous language models such as SciBERT \cite{beltagy2019scibert}, BioBERT \cite{lee2020biobert} and AlBERT \cite{lan2019albert} when analyzing twitter texts.


In order to study and extract information from social media texts it is neccessary to have a language model specific to social media domain. Futhermore, the TweetBERT models have been evaluated on 31 different datasets, including datasets from general, biomedical, scientific and Twitter domains. These state-of-the-art language representation models have shown promising results in the datasets for conducting text analysis. To show the effectiveness of our approach in Twitter text analysis, TweetBERTs were fine-tuned on two main Twitter text mining tasks: sentiment analysis and classification. In this paper, the authors made the following contribution:
\begin{itemize}
    \item TweetBERT, a domain specific language representation model trained on Twitter corpora for general Twitter text mining, is introduced.
    \item TweetBERT is evaluated on various Twitter datasets and is shown that both TweetBERTv1 and TweetBERTv2 outperform other traditional BERT models, such as BioBERT, SciBERT and BERT itself in Twitter text analysis.
    \item A comprehensive and elaborate analysis is provided by evaluating seven different BERT models including TweetBERTs on 31 different datasets, and their results are compared.
    \item Pre-trained weights of TweetBERT are released and source code is made available to the public \footnote{https://github.com/mohiuddin02/TweetBERT}.
\end{itemize}

The structure of the paper is as follows: The existing work in the field of language models is discussed in Section \ref{related}. Section \ref{methodology} presents the methodology, where it is described how the data has been collected for pre-training the model, and includes the approaches that were taken for implementing the TweetBERT models. In Section \ref{datasets} the datasets used for evaluating the model are described in detail. Section \ref{results} provides a discussion of the experimental results in the benchmark datasets with the various BERT and TweetBERT models. Finally, the conclusion is presented in Section \ref{conclusion}. 
\section{Related works} \label{related}
Recently a vast amount of work has been done, in the field of NLP, using bidirectional language models especially by modifying BERT  \cite{QudarSurvey}. BERT is a pre-trained neural network word representation model. It uses bidirectional transformer, which considers the sequence of data and, therefore, can understand the context of a text. It was pre-trained using texts from BookCorpus \cite{williams2017broad} and English Wiki \cite{devlin2018bert}. BERT uses two techniques for pre-training: masked language model, and next sentence prediction. Masking is carried out in three different ways in a sentence: by replacing a word with a token, or by replacing the word with a random word, or keeping the sentence as it is. These three ways help a bidirectional model to maintain and learn the context of a text. On the other hand, the next sentence prediction helps BERT to relate and connect two sentences together \cite{radford2018improving}. This is useful when evaluating sentiment analysis or question answering datasets. However, as BERT has been pre-trained on general corpora, it performs poorly in domain specific tasks. As a result, language models like BioBERT and SciBERT have been introduced. Recent language models have been broken down into two categories: contiual pre-training and pre-training from scratch.
\subsection{Continual Pre-training}
Continual models are those which use weights from another model and modify themselves for a specific task \cite{yoon2019collabonet}. BioBERT is a continual  pre-trained model because it was first initialized with the weights of BERT, and then pre-trained on various biomedical corpora, such as PubMed abstracts and PMC full-text articles, to make it domain specific \cite{lee2020biobert}. BioBERT was released as Biomedical documents were increasing and biomedical text analysis was becoming popular \cite{mnih2009scalable}. For example, more than 2,000 articles are published in biomedical peer-reviewed journals every day \cite{dougan2011context}. Directly using BERT to evaluate biomedical tasks did not give satisfactory results, thus BioBERT was created \cite{lee2020biobert}. BioBERT has the same architecture as BERT, but it has shown to perform better than BERT on biomedical text analysis \cite{Cho}. BioBERT was mainly evaluated in three biomedical tasks: biomedical named entity recognition, biomedical relation extraction, and biomedical question answering \cite{yoon2019collabonet}. Likewise, more models were introduced for specific domains. Lately, Covid-Twitter BERT model (CT-BERT) has been released to analyze tweets related to Covid \cite{ctbertmuller}. CT-BERT has been is pre-trained on around 160 million coronavirus tweets collected from the Crowdbreaks platform \cite{ctbertmuller}. CT-BERT is a continual BERT model and has shown an improvement of more than $10\%$ on classification datasets compared to the original BERT \cite{pires2019multilingual} model. This model has shown the most improvement in the target coronavirus related tweets. Furthermore, other extensions of BERT models, such as the AlBERT \cite{lan2019albert}, were also released. Generally, increasing the training corpus increases the performance of the NLP tasks. Moreover, the model size is directly proportional to the size of the training corpus. However, as the model size increases, it becomes increasely difficult to pre-train the model because there are GPU limitations. To address this factor AlBERT was introduced. It uses two parameter-reduction techniques to significantly reduces the number of training parameters in BERT: factorized embedding parameterization \cite{lan2019albert}, which breaks a large matrix into smaller matrices \cite{QudarSurvey}, and performing cross-layer parameter sharing, which cuts down the number of parameters as the neural network size increases. These methods have helped BERT to increase its training speed \cite{QudarSurvey}.

\subsection{Pre-training from Scratch}
There are other domains where both BERT and BioBERT provide unsatisfactory results. For example, when extracting information from general scientific texts, BERT performed poorly because it was only pre-trained on general domain corpora. As a result, SciBERT was released to evaluate scientific datasets \cite{beltagy2019scibert}. SciBERT also has the same architecture as BERT, but it is not a continual model. SciBERT is pre-trained from scratch and it uses a different WordPiece vocabulary called SentencePiece \cite{li2016biocreative}. SentencePiece vocabulary consists of words that are commonly used in scientific domains \cite{ammar2018construction}. When WordPiece and SentencePiece are compared, it is found that there is a similarity of only about 40\%. This shows that there is a huge difference between the words regularly used in general and scientific articles. SciBERT was pre-trained on a corpus from semantic scholar, containing 1.14 million papers from the computer science and biomedical domain \cite{gardner2018allennlp}. Each paper produced around 3,000 tokens making it similar to the number of tokens used to pre-train BERT \cite{gardner2018allennlp}. Similarly another BERT model was released, called RoBERTa, that showed that changing hyperparameter during pre-training BERT significantly increased the model’s performance \cite{liu2019roberta}. RoBERTa is not a continual model. It has been pre-trained on an extremely large, five different types of  corpora: BookCorpus, English Wikipedia, CC-News (collected from CommonCrawl News) dataset, OpenWebText, a WebText corpus \cite{radford2019language}, and Stories, a dataset containing story-like content \cite{radford2019language}. The overall size of the datasets was more than 160GB \cite{liu2019roberta}. Moreover, RoBERTa uses 4 different techniques, unlike BERT, to pre-train. They are: 
\begin{itemize}
    \item Segment-pair with next sentence prediction hyperparameter, which is the same as next sentence prediction as BERT \cite{liu2019roberta}.
    \item Sentence-pair next sentence prediction hyperparameter, where back to back sentences from only one document are connected \cite{liu2019roberta}.
    \item Full-sentences hyperparameter, where sentences within a document are connected.
    \item Doc-sentences hyperparameter, which is similar to full-sentences but two or more documents are connected \cite{radford2019language}.
\end{itemize}

\section{Methodology} \label{methodology}
This section discusses in detail the source of data collecting, tweet extracting, and corpora used for pre-training TweetBERT. An overview of the pre-training approach is shown in Fig. \ref{model}. There are two TweetBERT models: TweetBERTv1 and TweetBERTv2. Each of these models are pre-trained using different approaches, but have the same architecture as BERT because it is continual pre-training model. Moreover, Table \ref{modelcopora} shows the different variation of corpora and vocabulary used to pre-train each BERT model. For example, SciBERT uses SciVocab vocabulary which contain words popular in the scientific domain.
\subsection{Data Collection}
For domain specific text mining tasks, language models like BioBERT were pre-trained on PubMed and PMC \cite{lee2020biobert}. Likewise, TweetBERT was pre-trained on English tweets. TweetBERTv1 was pre-trained on a corpus that consists of 140 million tweets. The corpus contains tweets from top 100 personalities\footnote{https://www.kaggle.com/parulpandey/100-mostfollowed-twitter-accounts-as-of-dec2019} and top 100 hashtags of Twitter \cite{steenburgh2009hubspot}. Top personalities are the group of people who have the highest number of followers, Twitter platform. TweetBERTv2 was pre-trained on a similar but larger corpus containing 540 million English tweets. Table \ref{modelcopora} shows the different combination of corpora and WordPiece vocabulary involved in training of BERT models.\\
To create the training datasets tweets were collected and cleaned from \textit{big data analytics platform\footnote{https://twitter.datalab.science/}} developed in DaTALab at Lakehead University, Canada \cite{mendhe2020scalable}. This platform allows users to extract millions of tweets by simply giving keywords as inputs. The authors generated two corpora: Corpus140 and Corpus540 which indicate corpora with 140 and 540 million tweets, respectively. Each consists of tweets from top trending hashtags and top 100 personalities \cite{steenburgh2009hubspot}. The reason behind generating the corpora with the top personalities, followed by millions of Twitter users, was to ensure that the tweets were taken from authentic profile, since Twitter contains many fake accounts and their tweets have no real meaning. Moreover, tweets from top hashtags were used to analyze the pattern and style of informal language used in the Twitter platform by the general users.

\begin{table}[!h]
\begin{center}
\begin{tabular}{|c|c|c|}
 \rowcolor[HTML]{654321}
    \color[HTML]{FFFFFF}Model & 
    \color[HTML]{FFFFFF}Corpora Used & \color[HTML]{FFFFFF}WordPiece Vocab \\
    BERT & English Wiki + BookCorpus & BaseVocab\\
    SciBERT & Scientific articles & SciVocab\\
    TweetBERTv1 & English Wiki + BookCorpus + Corpus140 & BaseVocab\\
    TweetBERTv2 & English Wiki + BookCorpus + Corpus540 & BaseVocab + SciVocab\\
    \hline
    
\end{tabular}

\caption{Shows the different variation of corpora and WordPiece vocabulary involved in BERT models.}
\label{modelcopora}
\end{center}
\end{table}

\subsection{TweetBERT}

\begin{figure}[!h]
    \centering
    \includegraphics[width=\textwidth]{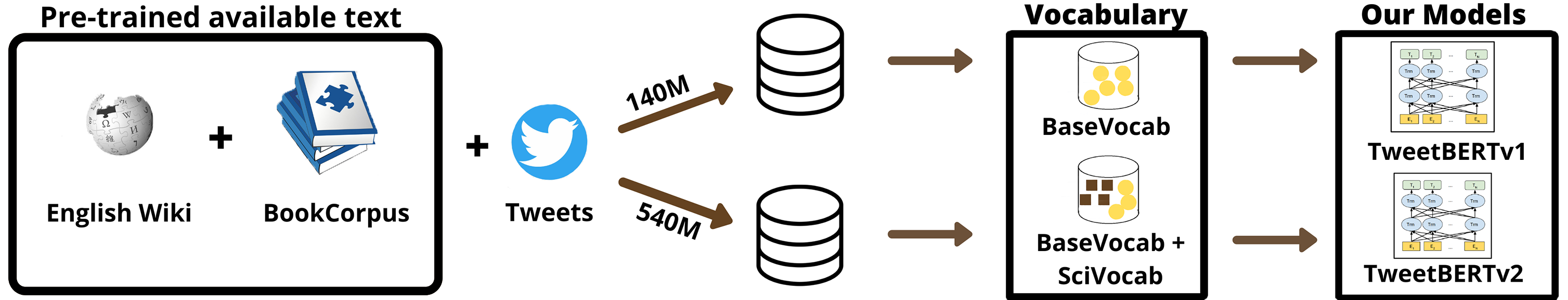}
    \caption{Overview of the pre-training TweetBERTs}
    \label{model}
\end{figure}
TweetBERT is a continual pre-trained model since it is initialized with the weight of BERT. BERT was pre-trained on general domain corpora that is English Wikipedia and BooksCorpus \cite{devlin2018bert}. As a result, the TweetBERT has the same architecture as BERT. Then, TweetBERTv1 is pre-trained on a Corpus140. On the other hand, TweetBERTv2 is initialzed with weights from AlBERT and pre-trained on Corpus540 using baseVocab and SciVocab. AlBERT model is the same as BERT, except that is uses two parameter-reduction techniques to reduce the number of training parameters in BERT, which increase the training speed of the model \cite{QudarSurvey} \cite{beltagy2019scibert}. BaseVocab and SciVocab are the WordPiece  and SentencePiece vocabulary of BERT and SciBERT, respectively. TweetBERTv1 uses the same WordPiece vocabulary as BERT so that the initial pre-trained weights of BERT and ALBERT are compatible with TweetBERT models \cite{lan2019albert}. TweetBERT, at the same time, uses the vocabulary of SciBERT so that scientific analysis can be carried out, for example detecting an epidemic or pandemic as opposed to simple sentiment analysis from tweets. TweetBERT can be also be used to evaluate other datasets in different domains, rather than just analyzing tweets. Fig. \ref{model} gives a detailed overview of the approach in making TweetBERT models.

\section{Datasets for Evaluation} \label{datasets}
In this article, results of evaluating seven BERT models on 31 different datasets are recorded. This section discusses some of the datasets that are used to evaluate the different BERT models including TweetBERTs. The datasets are divided into four domains: general, biomedical, scientific and Twitter domains. Each domain has its own set of datasets.
\subsection{General}
The general domain contains datasets such as GLUE \cite{wang2018glue}, SQuAD \cite{rajpurkar2016squad}, SWAG \cite{zellers2018swag} and RACE datasets. These datasets have contents that covers a wide range of general knowledge in basic English.
\subsubsection{GLUE}
The General Language Understanding Evaluation benchmark (GLUE) consists of datasets used for ``training, evaluating, and analyzing" language models \cite{wang2018glue}. GLUE consist of nine different datasets designed in such a way so that it can evaluate a model's understanding of general language\cite{QudarSurvey}\cite{warstadt2019neural}.
\begin{itemize}
    \item  Corpus of Linguistic Acceptability (CoLA) is a single-sentence task consisting of more than 10,000 English sentences. Each sentence is given a label indicating if its grammatical or ungrammatical English sentence. The language model's task is to predict the label \cite{QudarSurvey}.
    \item The Stanford Sentiment Treebank (SST) is also a binary single-sentence classification task containing sentences from movie reviews, along with their sentiment, labeled by humans \cite{socher2013recursive}. The task of language model is to predict the sentiment of a given sentence only.
    \item The Microsoft Research Paraphrase Corpus (MRPC) is a sentence pair corpus generated from online news sources, with human annotations for whether both the sentences are semantically equivalent or not. Thus, the task is to predict if a given sentence-pair has semantic similarity or not \cite{wang2018glue}.
    \item The Quora Question Pair(QQP) is similar to MRPC; the task is to predict how similar a given pair of questions are in terms of semantic meaning \cite{wang2018glue}. However, unlike MRPC, QQP dataset is a collection of questions from the question-answering website Quora\footnote{https://www.quora.com/} \cite{QudarSurvey}.
    \item The Semantic Textual Similarity Benchmark (STS-B) is a collections of sentence pairs extracted from news headlines, video and image captions, and similar sources, where semantic similarity score from one to five is assigned to the sentence pairs. The task is to predict the scores \cite{yang2018learning}. 
    \item The Multi-Genre Natural Language Inference Corpus (MNLI) is a crowd sourced dataset, consisting of sentence pairs with a human annotated premise and a hypothesis sentence. The task is to predict whether the premise sentence ``entails" the hypothesis, contradicts the hypothesis sentence or stays neutral \cite{wang2018glue}.
    \item Question Natural Language Inference (QNLI) is a simplified version of SQuAD dataset which has been converted into a binary classification task by forming a pair between each question and each sentence in the corresponding context \cite{QudarSurvey}. A language model's task would be to determine if the sentence contains the answer to the question. A positive value is assigned if pairs contain the correct answer, similarly a negative value is assigned if the pairs do not contain the answer \cite{wang2018glue}.
    \item Recognizing Textual Entailment (RTE) is similar to MNLI, where the language model predicts if a given sentence is similar to the hypothesis, contradicts or stays neutral. RTE dataset is very small compared to MNLI \cite{socher2013recursive}.
    \item The Winograd Schema Challenge (WNLI) is a reading comprehension task, in which a model takes a sentence with a pronoun as an input, and selects an answer from a list of choices that references to the given pronoun \cite{warstadt2019neural}.
\end{itemize}

\subsubsection{SQuAD}
The Stanford Question Answering Dataset is a collection of more than 100,000 questions answered by crowdworkers \cite{rajpurkar2016squad}. It contains 107,785 question-answer pairs on 536 articles \cite{QudarSurvey}. Each question and its following answer is from Wikipedia. SQuAD, unlike previous datasets like MCTest dataset \cite{richardson2013mctest}, does not provide a list of choices. The dataset has been created in such a way so that a language model can select the answer from the context of the passage and the question. In the beginning when releasing this dataset, logistic regression was performed to evaluate the level of difficultly \cite{pennington2014glove}. It was seen that the performance of the model decreases as the diversity of the model increases. The dataset helps a model to predict the context of a language \cite{richardson2013mctest}. 

\begin{tcolorbox}[enhanced jigsaw,breakable,pad at break*=1mm,
  colback=white,colframe=black,
  ]
  \textbf{Passage:} Apollo ran from 1961 to 1972, and was supported by the two-man Gemini program which ran concurrently with it from 1962 to 1966. Gemini missions developed some of the space travel techniques that were necessary for the success of the Apollo missions. Apollo used Saturn family rockets as launch vehicles. Apollo/Saturn vehicles were also used for an Apollo Applications Program, which consisted of Skylab, a space station that supported three manned missions in 1973–74, and the Apollo–Soyuz Test Project, a joint Earth orbit mission with the Soviet Union in 1975.\\

\textbf{Question:}\\
What space station supported three manned missions in 1973-1974\\

\textbf{Answer:}\\
Skylab\\

\end{tcolorbox}
\label{squad}
Fig. \ref{squad} is a sample from SQuAD dataset \cite{rajpurkar2016squad}.

\subsubsection{RACE}
Large-scale ReAding Comprehension Dataset From Examinations is a collection of approximately 28,000 English passages and 100,000 questions \cite{lai2017race}. This dataset was developed by English language professionals in a such a way so that a language model can gain an ability to read a passage or paragraph. The dataset is a multiple question answering task, where the model tries to predict the correct answer \cite{wang2018multi}\cite{habibi2017deep}\cite{QudarSurvey}. Other existing question answering dataset have two significant limitations. First, answer from any dataset can be found by simply a word-based search from the passage, which shows that the model is not able to consider the reasoning factor; this restricts the various types of questions that can be asked. Secondly, most datasets are crowd sourced \cite{QudarSurvey}, which introduces unwanted noise and bias in the dataset. Moreover, RACE is the largest dataset that support neural network training and needs logical reasoning to answer. It also contains option for an answer that might not be present in the training passage, which diversifies the questions that can be asked \cite{mccann2017learned}. RACE also contains content from various fields, allowing the language models to be more generic.
\subsubsection{SWAG }
Large-Scale Adversarial Dataset for Grounded Commonsense Inference dataset is composed of approximately 113,000 multiple choice questions, including 73,000 instances for training, 20,000 instances for validating, and 20,000 instances for testing, respectively \cite{zellers2018swag}. The multiple choice questions are derived from video caption, that are taken from ActivityNet Captions and the Large Scale Movie Description Challenge (LSMDC) \cite{shetty2015video}. The ActivityNet Captions consists of around 20,000 YouTube clips, in which each clip contains one of 203 activity types such as doing gymnastics or playing guitar \cite{bowman2015large}. LSMDC dataset has approximately 128,000 movie captions including both audio descriptions and scripts. For every captions pairs, constituency parsers have been used for splitting  the second sentence of each pair into nouns and verb phrases \cite{zellers2018swag}. Each question from the multiple choice questions was annotated by workers from Amazon Mechanical Turk. In order to improve the quality of the dataset, annotation artifacts were minimized. Annotation artifacts are the stylistic patterns that unintentionally provide suggestions for the target labels.

\subsection{Biomedical}
The biomedical domain contain datasets, such as NCBI, BC5CDR and MedNLI dataset. These datasets only contain texts related to biomedical domain.
\subsubsection{NCBI }
The national center for biotechnology information disease corpus is a collection of 793 PubMed abstracts in which abstracts are manually labelled by annotators, where the name of each disease and their corresponding concepts can be found in Medical Subject Headings \cite{habibi2017deep} or in Online Mendelian Inheritance in Man \cite{chiu2016named}. Name entity recognition is considered to be an important and challenging task of NLP. For example, \textit{adenomatous polyposis coli} \cite{yoon2019collabonet} and \textit{Friedrich ataxia} \cite{yoon2019collabonet} can be both a gene or a disease name. Also, abbreviated disease names are commonly used in biomedical texts, such as AS can stand for \textit{Angelanguage modelan syndrome, ankylosing spondylitis,aortic stenosis, Asperger syndrome} or even \textit{autism spectrum} \cite{wiese2017neural}\cite{QudarSurvey}. Also, doctors have their own way of describing a disease and as a result, it more difficult for any language model to achieve good performance. Evaluating a model on this NCBI dataset would show how the model performs in terms of remembering names, especially in biomedical domains \cite{cer2017semeval}.

\subsubsection{BC5CDR }
BC5CDR dataset consists of chemical induced disease (CID) relation extractions \cite{peng2019transfer}. The corpus is composed of 1,500 PubMed articles with approximately 4,400 annotated chemicals, 5,818 diseases and 3,116 chemical-disease interactions. To study the chemical interactions within diseases in depth, it is also not only important for the corpus to have the annotations of the chemical/diseases, but also their interactions with one another \cite{lee2020biobert}. Moreover, the corpus consists of disease/chemical annotations and relation annotations from the corresponding series of articles. Medical Subject Headings (MeSH) indexers were used for annotating the chemical/disease entities. Comparative Toxicogenomics Database (CTD) was used for annotating the CID relations. In order to attain a rich quality of annotation, comprehensive guidelines along with automatic annotation tools were given. For evaluating the inter-annotator agreement (IAA) score between each of the annotators, Jaccard similarity coefficient was calculated separately for the diseases and chemicals. This dataset has been used in multiple BioCreative V challenge assignments of biomedical text mining.\\
Chemical Disease Relations (CDR) are usually physically curated with the aid of CTD. However, this approach of curating manually is expensive. Thus, multiple alternative approaches have been proposed of guiding curation with text-mining mechanisms, which consist of the automatic extraction of CDRs. However, these proposed approaches have not been significantly successful since there are shortages of large training corpora. Moreover, to study the chemical interactions within diseases in depth, it is also not only important for the corpus to have the annotations of the chemicals diseases, but also their interactions with one another. However, there are multiple biomedical corpora that consist of only a few selected diseases and chemicals. In addition, none of the previous corpora have the instances of chemical-disease relation annotations, which includes abstracts having the entire chemical disease, relation annotation, and controlled vocabulary. In the case of the BC5CDR dataset, MeSH vocabulary was used as a controlled vocabulary similar to the existing biomedical information extraction datasets, BC5CDR that includes protein-protein interaction and drug-drug interactions. In contrast to the existing biomedical corpora, BC5CDR dataset is crucially different in terms of annotations (CID relations) from the 1,500 PubMed abstracts.

\subsubsection{MedNLI Dataset}
MedNLI dataset is a dataset that consists of medical history of the patients which is annotated by doctors. MIMIC-III have been used as the source of sentences. In order to avoid in annotating the data, only the medical prescriptions of deceased patients were used. The doctors performed a natural language inference task (NLI) task on the clinical notes that were provided. The MedNLI dataset has shown to be very handy as it is extremely challenging in having constructive, knowledge specific domains, where there is a shortage of training data. The clinical domain has a shortage of  massive-scale annotated datasets for training  machine learning models for natural language tasks, such as question answering, or paraphrasing. This makes the MedNLI a suitable resource in the open-medical field, since it is publicly available. Moreover, designing such a knowledge intensive medical domain dataset is expensive as well, since common approaches such as crowdsourcing platforms cannot be used for annotating the dataset. This is because annotating the dataset requires medical domain experts and thus curating such a dataset is very costly. Previously existing datasets have small sizes, and they target general fundamental natural language tasks such as co-reference resolution or information extraction tasks (e.g. named entity extraction).

\subsubsection{BIOSSES }
Biosses is one of the benchmark dataset for sentence similarity in the biomedical domain. The dataset is composed of 100 pairs of sentences. The sentences are selected from the Text Analysis Conference (TAC) containing Biomedical Summarization Track Training dataset. The TAC dataset consists of 20 reference articles and for each of the reference articles \cite{souganciouglu2017biosses}. The sentence pairs are mainly selected from the citing articles in which the sentence has a citation from any one of the reference articles. The data in TAC dataset is both semantically related. At the same time there are dissimilar sentence pairs that also occur in the annotated texts. Sentences that are citing articles from the same reference article will tend to be somewhat semantically similar \cite{chen2019biosentvec}. In addition, there are other sentences in which the citing sentence referring to an article is written about different ranges of topics or domains. Such sentence pairs will tend to have less or no similarity at all. Thus, sentence pairs covering different rates of similarity were obtained from the TAC dataset. In order to obtain a higher quality of dataset, only the pairs which gave strong alliance between the scores of the annotators were taken into account \cite{souganciouglu2017biosses}. Table \ref{Biosses} shows a sample from the original biosses dataset.

\begin{table}[!h]
    \centering
    \begin{tabular}{| m{5.5cm} | m{5.5cm}| m{5cm} |m{1cm} |}  \hline
 \rowcolor[HTML]{654321}
    \color[HTML]{FFFFFF}Sentence1 &
    \color[HTML]{FFFFFF}Sentence2 &\color[HTML]{FFFFFF}Comment & \color[HTML]{FFFFFF}Score\\
 
 \hline\hline
 Membrane proteins are proteins that interact with biological membranes. & Previous studies have demonstrated that membrane proteins are implicated in many diseases because they are positioned at the apex of signaling pathways that regulate cellular processes & The two sentences are not equivalent, but are on the same topic & 1 \\ 
 \hline
 This article discusses the current data on
using anti-HER2 therapies to treat CNS
metastasis as well as the newer anti-HER2
agents & Breast cancers with HER2 amplification
have a higher risk of CNS metastasis and
poorer prognosis & The two sentences are not equivalent, but
share some details & 2 \\
 \hline
 We were able to confirm that the cancer tissues had reduced expression of miR-126
and miR-424, and increased expression of
miR-15b, miR-16, miR-146a, miR-155
and miR-223 & A recent study showed that the expression of
miR-126 and miR-424 had reduced by
the cancer tissues & The two sentences are roughly equivalent,
but some important information differs/
missing & 3 \\
 \hline
 Hydrolysis of b-lactam antibiotics by b-lactamases is the most common mechanism of resistance for this class of antibacterial
agents in clinically important Gram-negative bacteria & In Gram-negative organisms, the most common b-lactam resistance mechanism involves b-lactamase-mediated hydrolysis
resulting in subsequent inactivation of the
antibiotic & The two sentences are completely or mostly
equivalent, as they mean the same thing & 4 \\
 \hline
\end{tabular}

\caption{A sample from Biosses dataset showing example annotations\cite{souganciouglu2017biosses}}
\label{Biosses}
\end{table}

\subsubsection{JNLPBA}
Joint Workshop on Natural language Processing in Biomedicine and its Application is a corpus of Pubmed abstracts specialized for NER tasks \cite{huang2019revised}. The types of entities that are selected from the biomedical domain include DNA, RNA, protein of cells and its types. However, few of the entities did not turn out to be prominently significant. For instance, entities for genes include the DNA as well as other gene entities like the protein and RNA \cite{huang2019revised}.

\subsubsection{Chemprot}
Chemprot is a chemical protein interaction corpus generated from PubMed abstracts \cite{taboureau2010chemprot}. The dataset consists of annotations within protein and chemical entities for identifying chemical protein interactions. The dataset is organized in a hierarchical structure with a total of 23 interactions. The author of the dataset has emphasized on mainly five high level interactions that includes: \textit{upregulator, downregulator, agonist, antagonist, and substrate} \cite{taboureau2010chemprot}.

\subsubsection{GAD}
Genetic Association Database is a dataset that was generated from the Genetic Association Archive \cite{jin2009gad}. The archive mainly contains gene-disease interactions from the sentences of PubMed abstracts. NER tool was also used in this dataset to detect gene-disease interactions and create artificial positive instances from the labeled archive sentences. On the other hand, negative instances from the dataset that were labeled as negative gene-disease interactions.

\subsubsection{HOC}
Hallmarks of Cancer dataset is generated from cancer hallmarks annotated on 1,499 PubMed abstracts. Afterwards, the dataset was broadened to 1,852 abstracts. The dataset has binary labels which focuses on labelling the cancer discussions on the abstracts as positive samples. However, the samples which had no mention of cancer were filtered out \cite{korn1997efficiently}.

\subsection{Scientific}
The scientific domain contain datasets such as SciCite and SCIERC that contain texts related to scientific domain.
\subsubsection{SCICITE}
SciCite is a dataset composed of citation intents that are extracted from various scientific fields \cite{beltagy2019scibert}. SciCite has been very recently released \cite{cohan2019structural}. The dataset was extracted from Semantic Scholar corpus of medical and computer science domains, and was annotated by giving label to citation content in four categories the are: \textit{method}, \textit{result}, \textit{comparison}, \textit{background}, and \textit{other}. Language models are used to evaluate how well it performs in classification and question answering tasks on scientific domain. 

\subsubsection{SCIERC}
SCIERC  dataset is a publicly available dataset that consists of annotations of around 5,000 scientific abstracts. The abstracts are collected from 12 AI conference/workshop proceedings from the Semantic Scholar Corpus. SCIERC is an extended version of previous existing similar datasets that are also collected from scientific articles, which include SemEval 2017 Task 10 \cite{cer2017semeval} and SemEval 2018 Task 7 \cite{cer2017semeval}. SCIERC  dataset is broadened in terms of summing up the cross-sentences related to one another by using conference links, named entity and relation types. 
\subsection{Twitter}
The Twitter domain contain datasets such as gender classification and tweets for sentiment analysis. These datasets only contain tweets.
\subsubsection{Twitter US airline dataset}
Twitter airline dataset\footnote{https://www.kaggle.com/crowdflower/twitter-airline-sentiment} is a collection of 14,640 tweets from six US airlines that includes: United, US Airways, Southwest, Delta and Virgin America. The tweets represent the reviews from each of the customers. The tweets are either labeled as positive, negative or neutral, based on the sentiment expressed. The airline company usually checks the feedback of their quality through traditional approaches such as the customer satisfaction questionnaires and surveys that are filled by customers. However, this approach is time consuming and inaccurate as customers might fill up the surveys in a hurry. Hence, designing an airline sentiment dataset as the Twitter airline dataset is very helpful since users in social media give genuine feedback and reviews about the airlines. 
\subsubsection{Twitter User Gender Classification}
In this dataset, \footnote{https://www.kaggle.com/crowdflower/twitter-user-gender-classification} annotators were asked to predict and label if the user of a certain Twitter account is male, female or a brand by only viewing the account. The dataset contains about 20,000 instances with user name, user id, account profile, account image and location.

\section{Results} \label{results}
In this Section the results of evaluating seven BERT models on 31 distinct datasets are discussed. The datasets used can be divided into four different domains. The general domain, includes eight datasets from GLUE \cite{wang2018glue}, SQuAD \cite{rajpurkar2016squad}, SWAG \cite{zellers2018swag} and RACE datasets. Table \ref{general} shows the performance of the BERT models on the general domain datasets. It is observed that ALBERT\cite{devlin2018bert} and RoBERTa \cite{liu2019roberta} achieve a higher score than other BERT models. AlBERT performs better in almost all of the GLUE datasets whereas RoBERTa outperforms in general question answering datasets. In Table \ref{general} the highest accuracies are underlined. The results of TweetBERT are fairly or sometimes extremely close to that of the highest accuracy. For example, on CoLA dataset AlBERT and TweetBERT achieves an accuracy of $71.42\%$ and $71\%$ respectively. Moreover, to understand the improvement and effectiveness of each TweetBERT models the marginal performance on each dataset is calculated using equation \ref{mp} \cite{ctbertmuller}. Table \ref{general_mp} shows the marginal performance between existing BERT models and TweetBERTv1 and Table \ref{table:general_mp_v2} shows the mariginal performance of TweetBERTv2 on general domain datasets. Positive value represents by how much the TweetBERT outperformed a BERT model. For example, from Table \ref{table:general_mp_v2} TweetBERT outperformed BioBERT by $12.81 \%$ in SQuAD dataset. On the other hand, negative value represents by how much the existing BERT model outperformed the TweetBERT model. To find the most suitable model overall on all the datasets the total of all the marginal performance of each BERT model was calculated. In the \textit{Total} row positive and negative number indicates the value by which TweetBERT performs better or worst than that BERT model. Both Table \ref{table:general_mp_v2} and \ref{general_mp} show that overall RoBERTa performs the best.
\begin{table}[!h]
\begin{tabular}{|l|l|l|l|l|l|l|l|l|l|l|}
\rowcolor[HTML]{654321} 
{\color[HTML]{FFFFFF} Domain} & {\color[HTML]{FFFFFF} Type} & {\color[HTML]{FFFFFF} Datasets} & {\color[HTML]{FFFFFF} Metrics} & {\color[HTML]{FFFFFF} BERT} & {\color[HTML]{FFFFFF} Biobert} & {\color[HTML]{FFFFFF} SciBERT} & {\color[HTML]{FFFFFF} RoBERTa} & {\color[HTML]{FFFFFF} Albert} & {\color[HTML]{FFFFFF} TweetBERT v1} & {\color[HTML]{FFFFFF} TweetBERT v2} \\
                              &                             & {\color[HTML]{000000} MNLI}     & A                              & 84.43                       & 86.27                          & 84.51                          & 90.28                          & 90.83                & \underline{90.91}                               & 90.51                               \\
                              &                             & {\color[HTML]{000000} QQP}      & A                              & 72.1                        & 85.65                          & 73.47                          & 92.21                & \underline{92.25}                         & 86.37                               & 88.83                               \\
                              &                             & {\color[HTML]{000000} QNLI}     & A                              & 90.51                       & 90.28                          & 88.34                          & 94.72                & \underline{95.37}                         & 91.25                               & 91.21                               \\
                              &                             & {\color[HTML]{000000} SST}      & A                              & 93.58                       & 93.86                          & 94.25                          & 96.4                  & \underline{96.99}                         & 92.43                               & 94.38                               \\
                              &                             & {\color[HTML]{000000} CoLA}     & A                              & 60.61                       & 65.83                          & 61.72                          & 68                             & \underline{71.42}                         & 68.42                               & 71                         \\
                              &                             & {\color[HTML]{000000} STS}      & PC                             & 86.51                       & 87.31                          & 87.14                          & 92.41                          & 96.94                         & 90.2                                & 94.41                      \\
                              &                             & {\color[HTML]{000000} MRPC}     & A                              & 89.3                        & 85.04                          & 90.78                          & 90.9                  & 90.9                & 88.64                               & \underline{91.79}                               \\
                              & \multirow{-8}{*}{GLUE}      & {\color[HTML]{000000} RTE}      & A                              & 70.11                       & 75.72                          & 66.26                          & 86.65                          & 89.21                & 75.23                               & \underline{91.3}                                \\ \cline{2-11} 
                              &                             & {\color[HTML]{000000} SQuad}    & A                              & 81.66                       & 72.22                          & 84.69                          & \underline{94.63}                          & 85.3                 & 69.84                               & 75.78                               \\
                              &                             & {\color[HTML]{000000} SWAG}     & A                              & 86.23                       & 82.71                          & 84.44                          & {\underline{90.16}}                    & 88.57                         & 85.47                               & 88.86                               \\
\multirow{-11}{*}{General}    & \multirow{-3}{*}{QA}        & {\color[HTML]{000000} RACE}     & A                              & 69.23                       & 80.9                           & 78.58                          & 81.31                          & \underline{82.37}                         & 81.96                               & 81.74 \\
\hline
\end{tabular}
\caption{Shows the performance of different BERT models on general domain dataset. Highest accuracies are underlined}
\label{general}
\end{table}

\begin{table}[!h]
\begin{tabular}{|c|c|c|r|r|r|r|r|}
\rowcolor[HTML]{654321} 
\color[HTML]{FFFFFF}Domain                             & \color[HTML]{FFFFFF}Type                  & \color[HTML]{FFFFFF}Datasets             & \multicolumn{1}{c}{\color[HTML]{FFFFFF}BERT} & \multicolumn{1}{c}{\color[HTML]{FFFFFF}Biobert} & \multicolumn{1}{c}{\color[HTML]{FFFFFF}SciBERT} & \multicolumn{1}{c}{\color[HTML]{FFFFFF}RoBERTa} & \multicolumn{1}{c}{\color[HTML]{FFFFFF}Albert} \\ \cline{1-8} 
\multirow{11}{*}{General}          & \multirow{8}{*}{GLUE} & MNLI                 & 41.61              & 33.79                & 41.31                & 6.48                & 0.87               \\
                                   &                       & QQP                  & 51.14              & 5.017                & 48.62                 & -74.96               & -75.87              \\
                                   &                       & QNLI                 & 7.79               & 9.97                 & 24.95                & -65.71                & -88.98             \\
                                   &                       & SST                  & -17.91             & -23.28                & -31.65                & -110.27              & -151.49              \\
                                   &                       & CoLA                 & 19.82              & 7.57                & 17.50                & 1.31                      & -10.49             \\
                                   &                       & STS                  & 27.35              & 22.77                 & 23.79                 & -29.11                & -220.26            \\
                                   &                       & MRPC                 & -6.16            & 24.06                & -23.21               & -24.83               & -24.83               \\
                                   &                       & RTE                  & 17.12             & -2.01               & 26.58                 & -85.54                & -129.56               \\ \cline{2-8} 
                                   & \multirow{3}{*}{QA}   & SQuAD                & -64.44             & -8.56                & -96.99                & -461.63                & -105.17               \\
                                   &                       & SWAG                 & -5.519            & 15.96                 & 6.61                 & -47.66                & -27.12                \\
                                   &                       & RACE                 & 20.77              & 5.54                  & 15.77                & 3.47                & -2.32               \\ \cline{1-8} 
\multicolumn{1}{|l}{\textbf{Total}} & \multicolumn{1}{l}{}  & \multicolumn{1}{l}{} & \textbf{91.59}     & \textbf{90.84}        & \textbf{53.32}        & \textbf{-888.49}       & \textbf{-835.25}     \\
\hline
\end{tabular}

\caption{Shows the marginal percentage of existing BERT models in comparison to TweetBERT v1 on different General datasets}
\label{general_mp}
\end{table}

\begin{table}[!h]
\begin{tabular}{|c|c|c|r|r|r|r|r|}
\rowcolor[HTML]{654321} 
\color[HTML]{FFFFFF}Domain                             & \color[HTML]{FFFFFF}Type                  & \color[HTML]{FFFFFF}Datasets             & \multicolumn{1}{c}{\color[HTML]{FFFFFF}BERT} & \multicolumn{1}{c}{\color[HTML]{FFFFFF}Biobert} & \multicolumn{1}{c}{\color[HTML]{FFFFFF}SciBERT} & \multicolumn{1}{c}{\color[HTML]{FFFFFF}RoBERTa} & \multicolumn{1}{c}{\color[HTML]{FFFFFF}Albert} \\ \cline{2-8} 
\multirow{11}{*}{General}           & \multirow{8}{*}{GLUE} & MNLI                 & 39.049            & 30.88                & 38.73                & 2.36                 & -3.48               \\
                                    &                       & QQP                  & 59.96              & 22.16                 & 57.89                 & -43.38                & -44.12               \\
                                    &                       & QNLI                 & 7.37             & 9.56                 & 24.61                 & -66.47                & -89.84               \\
                                    &                       & SST                  & 12.46              & 8.46                 & 2.26                 & -56.11                & -86.71               \\
                                    &                       & CoLA                 & 26.37              & 15.13                  & 24.24                 & 9.37                       & -1.46               \\
                                    &                       & STS                  & 58.56               & 55.94                 & 56.53                  & 26.35                 & -82.67               \\
                                    &                       & MRPC                 & 23.27              & 45.12                 & 10.95                 & 9.78                  & 9.78                 \\
                                    &                       & RTE                  & 70.89              & 64.16                 & 74.21                  & 34.83                 & 19.36              \\ \cline{2-8} 
                                    & \multirow{3}{*}{QA}   & SQuAD                & -32.06              & 12.81                  & -58.19                & -351.0              & -64.76               \\
                                    &                       & SWAG                 & 19.09              & 35.56                 &  45.1        & -13.21                & 2.53                \\
                                    &                       & RACE                 & 19.80              & 4.39                 & 14.75                 & 2.30                 & -3.57              \\ \hline
\multicolumn{1}{|l}{\textbf{Total}} & \multicolumn{1}{l}{}  & \multicolumn{1}{l}{} & \textbf{304.799}     & \textbf{304.22}        & \textbf{246.00}         & \textbf{-445.20}       & \textbf{-344.97}     \\
\hline
\end{tabular}

\caption{Shows the marginal percentage of existing BERT models in comparison to TweetBERT v2 on different General datasets}
\label{table:general_mp_v2}
\end{table}

\begin{equation} \label{mp}
\Delta MP = \frac{Accuracy_{BERT model} - Accuracy_{TweetBERTs}}{100 - Accuracy_{BERT model}} \times 100
\end{equation}

Secondly, the evaluation of the BERT models on 12 different biomedical domain datasets is shown in Table \ref{biomedical}. Precision, recall, and f1 score are used as metrics for measuring performance. It shows that, although BioBERT was pre-trained on millions of biomedical corpus, RoBERTa and TweetBERT outperforms BioBERT in all dataset types including NER and relation extraction. TweetBERTs performed best or very close to the best in many of the biomedical datasets. The the marginal performance of all the biomedical datasets between existing BERT models and TweetBERTs were calculate and reported in Table \ref{biomedical_mp} and \ref{biomedical_mp_v2} respectively. Results in both the table indicates that TweetBERT outperforms BERT, BioBERT and SciBERT.
\begin{table}[!h]
\begin{tabular}{|c|c|c|c|c|c|c|c|c|c|c|}
\rowcolor[HTML]{654321} 
\multicolumn{1}{l}{\color[HTML]{FFFFFF} Domain} & \multicolumn{1}{l}{\color[HTML]{FFFFFF} Type} & \multicolumn{1}{l}{\color[HTML]{FFFFFF} Datasets} & \multicolumn{1}{l}{\color[HTML]{FFFFFF} Metrics} & \multicolumn{1}{l}{\color[HTML]{FFFFFF} BERT} & \multicolumn{1}{l}{\color[HTML]{FFFFFF} Biobert} & \multicolumn{1}{l}{\color[HTML]{FFFFFF} SciBERT} & \multicolumn{1}{l}{\color[HTML]{FFFFFF} RoBERTa} & \multicolumn{1}{l}{\color[HTML]{FFFFFF} Albert} & \multicolumn{1}{l}{\color[HTML]{FFFFFF} TweetBERT v1} & \multicolumn{1}{l}{\color[HTML]{FFFFFF} TweetBERT v2} \\
                                                                          &                                                                         & {\color[HTML]{000000} NCBI}                                         & \begin{tabular}[c]{@{}c@{}}P\\ R\\ F\end{tabular}                          & \begin{tabular}[c]{@{}c@{}}88.30\\ 89.00\\ 88.60\end{tabular}           & \begin{tabular}[c]{@{}c@{}}88.22\\ 91.25\\ 89.71\end{tabular}              & 88.57                                                                      & \begin{tabular}[c]{@{}c@{}}\underline{90.97}\\ 91.15\\ \underline{90.58}\end{tabular}              & \begin{tabular}[c]{@{}c@{}}90.43\\ 91.22\\ 89.83\end{tabular}             & \begin{tabular}[c]{@{}c@{}}87.62\\ 91.33\\ 89.70\end{tabular}                   & \begin{tabular}[c]{@{}c@{}}90.38\\ \underline{91.62}\\ 89.69\end{tabular}                   \\ \cline{4-11} 
                                                                          &                                                                         & {\color[HTML]{000000} BC5CDR}                                       & \begin{tabular}[c]{@{}c@{}}P\\ R\\ F\end{tabular}                          &                                                     & \begin{tabular}[c]{@{}c@{}}86.47\\ 87.84\\ 87.15\end{tabular}              & 90.01                                                                      & \begin{tabular}[c]{@{}c@{}}\underline{90.28}\\ \underline{89.12}\\ \underline{90.64}\end{tabular}              & \begin{tabular}[c]{@{}c@{}}90.69\\ 89.03\\ 89.51\end{tabular}             & \begin{tabular}[c]{@{}c@{}}89.61\\ 86.09\\ 87.83\end{tabular}                   & \begin{tabular}[c]{@{}c@{}}89.22\\ 88.86\\ 90.41\end{tabular}                   \\ \cline{4-11} 
                                                                          &                                                                         & {\color[HTML]{000000} Species}                                              & \begin{tabular}[c]{@{}c@{}}P\\ R\\ F\end{tabular}                          & \begin{tabular}[c]{@{}c@{}}69.35\\ 74.05\\ 71.63\end{tabular}           & \begin{tabular}[c]{@{}c@{}}72.80\\ 75.36\\ 74.06\end{tabular}              & \begin{tabular}[c]{@{}c@{}}70.89\\ 75.82\\ 73.68\end{tabular}              & \begin{tabular}[c]{@{}c@{}}84.25\\ 87.16\\ 84.76\end{tabular}              & \begin{tabular}[c]{@{}c@{}}83.77\\ 85.90\\ 84.06\end{tabular}             & \begin{tabular}[c]{@{}c@{}}\underline{85.18}\\ 87.45\\ \underline{84.89}\end{tabular}                   & \begin{tabular}[c]{@{}c@{}}85.17\\ \underline{88.31}\\ 83.53\end{tabular}                   \\ \cline{4-11} 
                                                                          &                                                                         & {\color[HTML]{000000} BC5CDR}                                      & A                                                                          & 91.5                                                                    & 93                                                                         & 93.46                                                                      & 93.73                                                                      & 93.14                                                                     & 92.4                                                                            & 92.83                                                                           \\ \cline{4-11} 
                                                                          & \multirow{-5}{*}{NER}                                                   & {\color[HTML]{000000} JNLPBA}                                               & A                                                                          & 74.23                                                                   & 77.54                                                                      & 75.63                                                                      & 78.23                                                                      & 78.33                                                                     & 81.63                                                                           & 81.61                                                                           \\ \cline{2-11} 
                                                                          &                                                                         & {\color[HTML]{000000} GAD}                                                  & \begin{tabular}[c]{@{}c@{}}P\\ R\\ F\end{tabular}                          & \begin{tabular}[c]{@{}c@{}}79.21\\ 89.25\\ 83.25\end{tabular}           & \begin{tabular}[c]{@{}c@{}}77.32\\ 82.68\\ 79.83\end{tabular}              & \begin{tabular}[c]{@{}c@{}}80.18\\ 88.51\\ 80.28\end{tabular}              & \begin{tabular}[c]{@{}c@{}}\underline{83.82}\\ 90.14\\ 82.78\end{tabular}              & \begin{tabular}[c]{@{}c@{}}83.41\\ 89.73\\ 82.01\end{tabular}             & \begin{tabular}[c]{@{}c@{}}78.18\\ 91.81\\ 84.45\end{tabular}                   & \begin{tabular}[c]{@{}c@{}}78.11\\ \underline{91.92}\\ 85.57\end{tabular}                   \\ \cline{4-11} 
                                                                          &                                                                         & \color[HTML]{000000} EUADR                       & \begin{tabular}[c]{@{}c@{}}P\\ R\\ F\end{tabular}                          & \begin{tabular}[c]{@{}c@{}}75.45\\ 96.55\\ 84.62\end{tabular}           & \begin{tabular}[c]{@{}c@{}}84.83\\ 90.81\\ 80.92\end{tabular}              & \begin{tabular}[c]{@{}c@{}}74.91\\ \underline{96.64}\\ \underline{85.41}\end{tabular}              & \begin{tabular}[c]{@{}c@{}}\underline{85.84}\\ 89.5\\ 85.24\end{tabular}               & \begin{tabular}[c]{@{}c@{}}85.76\\ 90.48\\ 84.11\end{tabular}             & \begin{tabular}[c]{@{}c@{}}77.73\\ 92.31\\ 81.36\end{tabular}                   & \begin{tabular}[c]{@{}c@{}}75.95\\ 92.1\\ 79.39\end{tabular}                    \\ \cline{4-11} 
                                                                          & \multirow{-3}{*}{RE}                                                    & {\color[HTML]{000000} CHEMPROT}                                             & \begin{tabular}[c]{@{}c@{}}P\\ R\\ F\end{tabular}                          & \begin{tabular}[c]{@{}c@{}}76.02\\ 71.60\\ 73.74\end{tabular}           & \begin{tabular}[c]{@{}c@{}}77.02\\ 75.90\\ 76.46\end{tabular}              & 71.3                                                                       & \begin{tabular}[c]{@{}c@{}}80.17\\ 78.97\\ 79.32\end{tabular}              & \begin{tabular}[c]{@{}c@{}}85.32\\ 87.55\\ 83.29\end{tabular}             & \begin{tabular}[c]{@{}c@{}}\underline{86.10}\\ 84.35\\ \underline{85.63}\end{tabular}                   & \begin{tabular}[c]{@{}c@{}}85.77\\ \underline{87.69}\\ 85.04\end{tabular}                   \\ \cline{2-11} 
                                                                          &                                                                         & {\color[HTML]{000000} MedSTS}                                               & A                                                                          & 78.6                                                                    & 84.5                                                                       & 78.6                                                                       & 89.06                                                                      & \underline{91.06}                                                                    & 86.78                                                                           & 90.89                                                                           \\ 
                                                                          & \multirow{-2}{*}{Sentence}                                   & {\color[HTML]{000000} Biosses}                                              & A                                                                          & 71.2                                                                    & 82.7                                                                       & 74.23                                                                      & 88.77                                                                      & \underline{91.25}                                                                     & 80.27                                                                           & 83.96                                                                           \\ \cline{2-11} 
                                                                          & Inference                                                               & {\color[HTML]{000000} MedLNI}                                               & A                                                                          & 75.4                                                                    & 80.5                                                                       & 75.36                                                                      & 86.39                                                                      & \underline{90.13}                                                                     & 82.16                                                                           & 88.41                                                                           \\
                                                                          \cline{2-11}
\multirow{-12}{*}{Biomedical}                                             & Doc classif                                                 & {\color[HTML]{000000} HoC}                                                  & A                                                                          & 80                                                                      & 82.9                                                                       & 80.12                                                                      & 87.83                                                                      & \underline{91.48}                                                                     & 82.71                                                                           & 86                                                                              \\  \cline{1-11} 
\end{tabular}

\caption{Shows the performance of different BERT models on biomedical domain dataset. Highest accuracies are underlined.}
\label{biomedical}
\end{table}

\begin{table}[!h]
\begin{tabular}{|c|c|c|r|r|r|r|r|}
\rowcolor[HTML]{654321} 
\color[HTML]{FFFFFF}Domain                             & \color[HTML]{FFFFFF}Type                  & \color[HTML]{FFFFFF}Datasets             & \multicolumn{1}{c}{\color[HTML]{FFFFFF}BERT} & \multicolumn{1}{c}{\color[HTML]{FFFFFF}Biobert} & \multicolumn{1}{c}{\color[HTML]{FFFFFF}SciBERT} & \multicolumn{1}{c}{\color[HTML]{FFFFFF}RoBERTa} & \multicolumn{1}{c}{\color[HTML]{FFFFFF}Albert} \\
                                   &                                       & {\color[HTML]{000000} NCBI disease}         & 9.64                     & 0.91                        & 9.88                        & -9.34                       & -1.27                      \\
                                   &                                       & {\color[HTML]{000000} BC5CDR disease}       &      & -0.14                       & -21.82                      & -30.02                      & -16.01                     \\
                                   &                                       & {\color[HTML]{000000} Species}              & 46.7                     & 49.06                       & 42.59                       & 0.85                        & 5.2                        \\
                                   &                                       & {\color[HTML]{000000} BC5CDR chemical}      & 10.58             & -8.57                & -16.20                & -21.21                & -10.78            \\
                                   & \multirow{-5}{*}{NER}                 & {\color[HTML]{000000} JNLPBA}               & 28.71             & 18.21                 & 24.62                 & 15.61                 & 15.22                 \\ \cline{2-8} 
                                   &                                       & {\color[HTML]{000000} GAD}                  & 7.16                     & 52.71                       & 21.14                       & 9.69                        & 13.56                      \\
                                   &                                       & {\color[HTML]{000000} EUADR}                & -21.19                   & 16.32                       & -27.75                      & -26.54                      & -17.3                      \\
                                   & \multirow{-3}{*}{RE}                  & {\color[HTML]{000000} CHEMPROT}             & 45.27                    & 35.06                       & 49.93                       & 30.51                       & 14                         \\ \cline{2-8} 
                                   &                                       & {\color[HTML]{000000} MedSTS}               & 38.22            & 14.70                & 38.22                 & -20.84                & -47.87              \\
                                   & \multirow{-2}{*}{Sent sim} & {\color[HTML]{000000} Biosses}              & 31.49              & -14.047                & 23.43                 & -75.69               & -125.48               \\ \cline{2-8} 
                                   & Inference                             & {\color[HTML]{000000} MedLNI}               & 27.47              & 8.51                & 27.59                  & -31.08                & -80.79               \\ \cline{2-8} 
\multirow{-12}{*}{Biomedical}      & Doc Classifi               & {\color[HTML]{000000} HoC}                  & 13.55                    & -1.11               & 13.02                 & -42.07                & -102.93              \\ \hline
\multicolumn{1}{|l}{\textbf{Total}} & \multicolumn{1}{l}{}                  & \multicolumn{1}{l}{{\color[HTML]{000000} }} & \textbf{237.63}     & \textbf{171.62}        & \textbf{184.67}        & \textbf{-200.12}       & \textbf{-354.42}     \\
\hline
\end{tabular}

\caption{Shows the marginal percentage of existing BERT models in comparison to TweetBERT v1 on different Biomedical datasets.}
\label{biomedical_mp}
\end{table}

\begin{table}[!h]
\begin{tabular}{|c|c|c|r|r|r|r|r|}
\rowcolor[HTML]{654321} 
\color[HTML]{FFFFFF}Domain                             & \color[HTML]{FFFFFF}Type                  & \color[HTML]{FFFFFF}Datasets             & \multicolumn{1}{c}{\color[HTML]{FFFFFF}BERT} & \multicolumn{1}{c}{\color[HTML]{FFFFFF}Biobert} & \multicolumn{1}{c}{\color[HTML]{FFFFFF}SciBERT} & \multicolumn{1}{c}{\color[HTML]{FFFFFF}RoBERTa} & \multicolumn{1}{c}{\color[HTML]{FFFFFF}Albert} \\
                                    &                          & {\color[HTML]{000000} NCBI disease}         & 9.56                     & -0.19                       & 9.79                        & -9.44                       & -1.37                      \\
                                    &                        & {\color[HTML]{000000} BC5CDR disease}       &    6.2    & 25.36                       & 4                           & -2.45                       & 8.57                       \\
                                    &                          & {\color[HTML]{000000} Species}              & 41.9                     & 36.5                        & 37.4                        & -8.07                       & -3.32                      \\
                                    &                          & {\color[HTML]{000000} BC5CDR chemical}      & 15.64             & -2.42               & -9.63                & -14.35                & -4.51               \\
                                    & \multirow{-5}{*}{NER}    & {\color[HTML]{000000} JNLPBA}               & 28.63              & 18.12                 & 24.53                 & 15.52                & 15.13                 \\ \cline{2-8} 
                                    &                          & {\color[HTML]{000000} GAD}                  & 13.85                    & 28.45                       & 29.67                       & 16.2                        & 19.78                      \\
                                    &                          & {\color[HTML]{000000} EUADR}                & -30.49                   & -5.18                       & -37.55                      & -35.98                      & -26.3                      \\
                                    & \multirow{-3}{*}{RE}     & {\color[HTML]{000000} CHEMPROT}             & 43.03                    & 36.44                       & 47.87                       & 27.65                       & 10.47                      \\ \cline{2-8} 
                                    &                          & {\color[HTML]{000000} MedSTS}               & 57.42              & 41.22                 & 57.42                 & 16.72                 & -1.90               \\
                                    & \multirow{-2}{*}{Sen sim} & {\color[HTML]{000000} Biosses}              & 44.30              & 7.28                 & 37.75                 & -42.83                 & -83.31              \\ \cline{2-8} 
                                    & Inference                & {\color[HTML]{000000} MedLNI}               & 52.88              & 40.56                 & 52.96                 & 14.84                 & -17.42              \\ \cline{2-8} 
\multirow{-12}{*}{Biomedical}       & Doc Classifi             & {\color[HTML]{000000} HoC}                  & 30                       & 18.12                 & 29.57                 & -15.03                & -64.31               \\ \hline
\multicolumn{1}{|l}{\textbf{Total}} & \multicolumn{1}{l}{}     & \multicolumn{1}{l}{{\color[HTML]{000000} }} & \textbf{306.75}     & \textbf{244.27}        & \textbf{283.81}        & \textbf{-37.21}       & \textbf{-148.51}     \\
\hline
\end{tabular}

\caption{Shows the marginal percentage of existing BERT models in comparison to TweetBERT v2 on different Biomedical datasets.}
\label{biomedical_mp_v2}
\end{table}

Thirdly, BERT models on four scientific datasets were evaluated. Previously, with the introduction of SciBERT there was statistical evidence that it performed remarkably better on scientific tasks. Although, Table \ref{sci} show that TweetBERT performed best in only two datasets Table \ref{sci_mp_v2} show that TweetBERTv2 outperformed SciBERT and it is more suitable to use TweetBERTv2 to evaluate scientific tasks rather than using SciBERT. One of the main reason of TweetBERTv2 performing better than SciBERT is because of using SciBERT's vocabulary when pre-training TweetBERTv2.

\begin{table}[!h]
\begin{tabular}{|c|c|c|c|c|c|c|c|c|c|c|}
\rowcolor[HTML]{654321} 
\hline
\multicolumn{1}{l}{{\color[HTML]{FFFFFF} Domain}}   & \multicolumn{1}{l}{{\color[HTML]{FFFFFF} Type}}              & \multicolumn{1}{l}{{\color[HTML]{FFFFFF} Datasets}} & \multicolumn{1}{l}{{\color[HTML]{FFFFFF} Metrics}} & \multicolumn{1}{l}{{\color[HTML]{FFFFFF} BERT}} & \multicolumn{1}{l}{{\color[HTML]{FFFFFF} Biobert}} & \multicolumn{1}{l}{{\color[HTML]{FFFFFF} SciBERT}} & \multicolumn{1}{l}{{\color[HTML]{FFFFFF} RoBERTa}} & \multicolumn{1}{l}{{\color[HTML]{FFFFFF} Albert}} & \multicolumn{1}{l}{{\color[HTML]{FFFFFF} TweetBERT v1}} & \multicolumn{1}{l}{{\color[HTML]{FFFFFF} TweetBERT v2}} \\ \hline
{\color[HTML]{000000} }                             & {\color[HTML]{000000} }                                      & {\color[HTML]{000000} Paper feild}                  & {\color[HTML]{000000} A}                           & {\color[HTML]{000000} 55.06}                    & {\color[HTML]{000000} 56.22}                       & {\color[HTML]{000000}  65.71}                 & {\color[HTML]{000000} 63.48}                       & {\color[HTML]{000000} 62.85}                      & {\color[HTML]{000000} 58.12}                            & {\color[HTML]{000000} \underline{ 66.49}}                            \\
{\color[HTML]{000000} }                             & {\color[HTML]{000000} }                                      & {\color[HTML]{000000} Sci-cite}                     & {\color[HTML]{000000} A}                           & {\color[HTML]{000000} 84.33}                    & {\color[HTML]{000000} 85.11}                       & {\color[HTML]{000000} 85.42}                       & {\color[HTML]{000000} 87.16}                       & {\color[HTML]{000000} 86.68}                      & {\color[HTML]{000000}  88.5}                       & {\color[HTML]{000000} \underline{ 88.56}}                            \\
{\color[HTML]{000000} }                             & \multirow{-3}{*}{{\color[HTML]{000000} Text Classi}} & {\color[HTML]{000000} Sci-RE}     & {\color[HTML]{000000} A}                           & {\color[HTML]{000000} 63.55}                    & {\color[HTML]{000000} 65.42}                       & {\color[HTML]{000000} 65.77}                       & {\color[HTML]{000000} 66.79}                       & {\color[HTML]{000000} 68.46}                      & {\color[HTML]{000000} \underline{ 68.85}}                      & {\color[HTML]{000000} 66.82}                            \\ \cline{2-11}
\multirow{-4}{*}{{\color[HTML]{000000} Scientific}} & {\color[HTML]{000000} Parsing}                               & {\color[HTML]{000000} Genia}                        & {\color[HTML]{000000} A}                           & {\color[HTML]{000000} 64.81}                    & {\color[HTML]{000000} 67.71}                       & {\color[HTML]{000000} 72.3}                        & {\color[HTML]{000000} 76.95}                       & {\color[HTML]{000000} \underline{ 78.45}}                & {\color[HTML]{000000} 67.98}                            & {\color[HTML]{000000} 70}                               \\
\hline
\end{tabular}

\caption{Shows the performance of different BERT models on scientific domain dataset. Highest accuracies are underlined.}
\label{sci}
\end{table}

\begin{table}[!h]
\begin{tabular}{|c|c|c|r|r|r|r|r|}
\rowcolor[HTML]{654321} 
\color[HTML]{FFFFFF}Domain                             & \color[HTML]{FFFFFF}Type                  & \color[HTML]{FFFFFF}Datasets             & \multicolumn{1}{c}{\color[HTML]{FFFFFF}BERT} & \multicolumn{1}{c}{\color[HTML]{FFFFFF}Biobert} & \multicolumn{1}{c}{\color[HTML]{FFFFFF}SciBERT} & \multicolumn{1}{c}{\color[HTML]{FFFFFF}RoBERTa} & \multicolumn{1}{c}{\color[HTML]{FFFFFF}Albert} \\
                                    &                                       & {\color[HTML]{000000} paper feild}           & 6.80              & 4.33                & -22.13               & -14.67                & -12.73               \\
                                    &                                       & sci-cite                & 26.61              & 22.76                 & 21.12                 & 10.43                 & 13.66                \\
                                    & \multirow{-3}{*}{Text Classi} &  sci-RE & 14.54              & 9.91                 & 8.99                 & 6.20                 & 1.23                \\ \cline{2-8} 
\multirow{-4}{*}{Scientific}        & Parsing                               & {\color[HTML]{000000} Genia}                 & 9.00              & 0.83                & -15.59                & -38.91                 & -48.58              \\ \hline
\multicolumn{1}{|l}{\textbf{Total}} & \multicolumn{1}{l}{}                  & \multicolumn{1}{l|}{}                        & \textbf{56.96}     & \textbf{37.86}        & \textbf{-7.60}       & \textbf{-36.95}       & \textbf{-46.41}     \\
\hline
\end{tabular}

\caption{Shows the marginal percentage of existing BERT models in comparison to TweetBERT v1 on different scientific datasets.}
\label{sci_mp}
\end{table}

\begin{table}[!h]
\begin{tabular}{|c|c|c|r|r|r|r|r|}
\rowcolor[HTML]{654321} 
\color[HTML]{FFFFFF}Domain                             & \color[HTML]{FFFFFF}Type                  & \color[HTML]{FFFFFF}Datasets             & \multicolumn{1}{c}{\color[HTML]{FFFFFF}BERT} & \multicolumn{1}{c}{\color[HTML]{FFFFFF}Biobert} & \multicolumn{1}{c}{\color[HTML]{FFFFFF}SciBERT} & \multicolumn{1}{c}{\color[HTML]{FFFFFF}RoBERTa} & \multicolumn{1}{c}{\color[HTML]{FFFFFF}Albert}\\
                                    &                                       & {\color[HTML]{000000} paper feild} & 25.43              & 23.45                & 2.27                & 8.24                & 9.79               \\
                                    &                                       & {\color[HTML]{333333} sci-cite}    & 26.99             & 23.16                 & 21.53                 & 10.90                & 14.11              \\
                                    & \multirow{-3}{*}{Text Classification} & {\color[HTML]{333333} sci-RE}      & 8.97             & 4.04                 & 3.06                & 0.09              & -5.19               \\ \cline{2-8} 
\multirow{-4}{*}{Scientific}        & Parsing                               & {\color[HTML]{000000} genia}       & 14.74              & 7.09               & -8.30                & -30.15                & -39.21             \\ \hline
\multicolumn{1}{|l}{\textbf{Total}} & \multicolumn{1}{l}{}                  & \multicolumn{1}{l|}{}              & \textbf{76.14}     & \textbf{57.76}        & \textbf{18.57}        & \textbf{-10.91}       & \textbf{-20.49}     \\
\hline
\end{tabular}

\caption{Shows the marginal percentage of existing BERT models in comparison to TweetBERT v2 on different scientific datasets.}
\label{sci_mp_v2}
\end{table}

Finally, all the BERT models were evaluated on tweets sentiment and classification datasets. As the TweetBERTs were pre-trained on millions of tweets it outperformed all existing BERT models, as expected. Table \ref{twitter} records the performance of the BERT models including our TweetBERT. Table \ref{twitter_mp} shows that the highest total marginal performance is $159.15\%$ when SciBERT and TweetBERTv1 are compared. Table \ref{twitter_mp_v2}, on the other hand, shows that the lowest marginal performance, $167.1\%$, is greater than the highest marginal performance from Table \ref{twitter_mp}. As a result, it can concluded that TweetBERTv2 performs significantly better than TweetBERTv1 in Twitter domain tasks.

\begin{table}[!h]
\begin{tabular}{|c|c|c|c|c|c|c|c|c|c|c|}
\rowcolor[HTML]{654321} 
\color[HTML]{FFFFFF} Domain                  & \color[HTML]{FFFFFF}Type                 &\color[HTML]{FFFFFF} Datasets                           & \color[HTML]{FFFFFF}\color[HTML]{FFFFFF}Metrics & \color[HTML]{FFFFFF}BERT  & \color[HTML]{FFFFFF}Biobert &\color[HTML]{FFFFFF} SciBERT &\color[HTML]{FFFFFF} RoBERTa & \color[HTML]{FFFFFF}Albert & \color[HTML]{FFFFFF}TweetBERTv1 & \color[HTML]{FFFFFF}TweetBERTv2 \\
\multirow{4}{*}{Twitter} & Sentiment   & Airline Senti      & A       & 85.2  & 84.17   & 82.73   & 88.68   & 87.08  & 89           & \underline{ 92.99}  \\
                         &      & Gender Classi & A       & 80.65 & 80.22   & 72.23   & 80.74   & 82.22  & 85.02        & \underline{ 89.75}  \\ 
                         &  & Sentiment140                       & A       & 85.63 & 87.84   & 82.29   & 86.71   & 90.59  & 92.74        & \underline{ 95.18}  \\
                         &  & Political Tweets                   & A       & 69.99 & 69.34   & 64.66   & 72.01   & 69.57  & 75.13        & \underline{ 78.79} \\ 
                         \hline
\end{tabular}

\caption{Shows the performance of BERT models in different Twitter datasets. Highest accuracies are underlined.}
\label{twitter}
\end{table}

\begin{table}[!h]
\begin{tabular}{|c|l|c|r|r|r|r|r|}
\rowcolor[HTML]{654321} 
\color[HTML]{FFFFFF}Domain                               & \multicolumn{1}{c|}{\color[HTML]{FFFFFF}Type}     & \color[HTML]{FFFFFF}Datasets              & \multicolumn{1}{c|}{\color[HTML]{FFFFFF}BERT} & \multicolumn{1}{c|}{\color[HTML]{FFFFFF}Biobert} & \multicolumn{1}{c|}{\color[HTML]{FFFFFF}SciBERT} & \multicolumn{1}{c|}{\color[HTML]{FFFFFF}RoBERTa} & \multicolumn{1}{c|}{\color[HTML]{FFFFFF}Albert} \\
\multirow{4}{*}{Twitter}             & \multirow{3}{*}{Sentiment} & Airline Senti       & 25.67             & 30.51                 & 36.30                 & 2.82                 & 14.86                               \\
                                    &                                     & Gender Classi & 22.58              & 24.26                & 34.21                & 22.22                 & 15.74                                \\
                                    &                                     & Sentiment140                       & 49.47              & 40.29                 & 59.0                 & 45.37                  & 22.84                               \\
                                    &                & Political Tweets                  & 17.12              & 18.88                 & 29.62                 & 11.14                & 18.27                              \\ \hline
\multicolumn{1}{|l}{\textbf{Total}} &   \multicolumn{1}{l}{}               & \multicolumn{1}{l}{}               & \textbf{114.86}     & \textbf{113.95}        & \textbf{159.15}        & \textbf{81.56}          & \multicolumn{1}{r|}{\textbf{71.72}} \\
\hline
\end{tabular}

\caption{Shows the marginal percentage of existing BERT models in comparison to TweetBERTv1 on different Twitter datasets.}
\label{twitter_mp}
\end{table}

\begin{table}[!h]
\begin{tabular}{|c|l|c|r|r|r|r|r|}
\rowcolor[HTML]{654321} 
\color[HTML]{FFFFFF}Domain                               & \multicolumn{1}{c|}{\color[HTML]{FFFFFF}Type}     & \color[HTML]{FFFFFF}Datasets              & \multicolumn{1}{c|}{\color[HTML]{FFFFFF}BERT} & \multicolumn{1}{c|}{\color[HTML]{FFFFFF}Biobert} & \multicolumn{1}{c|}{\color[HTML]{FFFFFF}SciBERT} & \multicolumn{1}{c|}{\color[HTML]{FFFFFF}RoBERTa} & \multicolumn{1}{c|}{\color[HTML]{FFFFFF}Albert}               \\
\multirow{4}{*}{Twitter}            & \multirow{3}{*}{Sentiment} &  Airline Senti      & 52.63              & 55.716                & 59.40                 & 38.07               & 45.74                              \\
                                    &                                     &  Gender Classi & 47.02              & 48.17                 & 54.98                 & 46.78                 & 42.35                             \\
                                    &                                     & Sentiment140                       & 66.45              & 60.36                & 72.78                  & 63.73                 & 48.77                             \\
                                    &                 & Political Tweets                   & 29.32             & 30.82                 & 39.98                 & 24.22                 & 30.29                             \\ \hline
\multicolumn{1}{|l}{\textbf{Total}} & \multicolumn{1}{l}{}                & \multicolumn{1}{l}{}               & \textbf{195.44}     & \textbf{195.08}        & \textbf{227.16}        & \textbf{172.81}        & \multicolumn{1}{r|}{\textbf{167.17}} \\
\hline
\end{tabular}
\caption{Shows the marginal percentage of existing BERT models in comparison to TweetBERTv2 on different Twitter datasets.}
\label{twitter_mp_v2}
\end{table}

\newpage

\section{Conclusion} \label{conclusion}
Twitter is a popular social networking site, which contain valuable data, where analyzing the content is particularly challenging. Tweets are usually written in an informal structure, and as a consequence, using language models trained on general domain corpora like BERT or other domains such as BioBERT often gives unsatisfactory results. Hence, two versions of TweetBERT are introduced, which are pre-trained language representation models used for Twitter text mining. This paper also discusses how the data was collected from the \textit{big data analytics platform} for pre-training TweetBERT. Millions of tweets were extracted and cleaned from this platform. Moreover, detailed discussion of pre-training TweetBERT models are included. TweetBERTv1 was initialized using weights from BERT and then pre-trained on a tweet corpus. In the case of TweetBERTv2, first the model is initialized with weights from AlBERT and used vocabularies from both BERT and SciBERT. Two main advantages of using BaseVocab and SciVocab are scientific analysis can be carried out by studying tweets, and ALBERT is compatible with TweetBERTs and can be used in other evaluating other datasets in different domains rather than just analyzing tweets.\\
Moreover, this paper focuses on the datasets used to evaluate BERT models. Evaluation of TweetBERT models and five other BERT models on 31 different datasets from general, biomedical, scientific and Twitter domains and provide a comparison between them. Section \ref{datasets} gives a detail description of most of the datasets used. Finally, the results for the evaluation are released. It is shown that TweetBERT significantly outperforms other BERT models on Twitter datasets, and even on some other domain datasets, like BioBERT. The marginal performance that shows the amount by which a BERT model outperforms another BERT model is calculate. It shows that, especially in the case of Twitter datasets, TweetBERTs has the best performance. TweetBERTv2 outperforms AlBERT by a total of 167.17\% when evaluating Twitter datasets. \\
Overall, an extensive discussion is provided about the necessity of language model specific to social media. We introduce TweetBERTs and give comprehensive discussion about the methods, approaches and data used to pre-train TweetBERTs. Moreover, a detailed analysis is carried out and released the results by evaluating seven different BERT models on 31 different datasets.

\section*{Acknowledgement}
This research was funded by Natural Sciences and Engineering Research Council (NSERC), Canada. The authors would also like to thank DaTALab researchers Dhivya, Zainab and Punar for providing their valuable inputs, Lakehead University's HPC for providing high-end GPUs and CASES building for providing the infrastructure.

\bibliographystyle{IEEEtran}
\bibliography{IEEEexample}

\end{document}